\documentclass[letterpaper]{article}
\usepackage{ijcai13}
\usepackage{times}
\usepackage{ijcai13}
\usepackage{helvet}
\usepackage{courier}
\usepackage{amsmath}
\usepackage{amsfonts}
\usepackage{bm} %mathbb的包
\usepackage{graphicx}
\usepackage{algorithm}
\usepackage{algorithmic}
\usepackage{caption}
\usepackage{bm}
\usepackage{color}
\usepackage{times}
\usepackage{multirow}
\title{Online Group Feature Selection}
\author{Jing Wang$^1$, Zhong-Qiu Zhao$^{1,2}$, Xuegang Hu$^1$,\\ \textbf{Yiu-ming Cheung$^2$, \textbf{Meng Wang$^1$,} Xindong Wu$^{1,3}$\thanks{Corresponding author.}} \\
$^1$ College of Computer Science and Information Engineering,\\
 Hefei University of Technology, Hefei 230009, China\\
$^1$hfutwj@gmail.com, $^3$xwu@uvm.edu\\
}
\begin{document}

\maketitle

\begin{abstract}
Online feature selection with dynamic features has become an active research area in recent years. However, in some real-world applications such as image analysis and email spam filtering, features may arrive by groups. Existing online feature selection methods evaluate features individually, while existing group feature selection methods cannot handle online processing. Motivated by this, we formulate the online group feature selection problem, and propose a novel selection approach for this problem. Our proposed approach consists of two stages: online intra-group selection and online inter-group selection. In the intra-group selection, we use spectral analysis to select discriminative features in each group when it arrives. In the inter-group selection, we use Lasso to select a globally optimal subset of features. This 2-stage procedure continues until there are no more features to come or some predefined stopping conditions are met. Extensive experiments conducted on benchmark and real-world data sets demonstrate that our proposed approach outperforms other state-of-the-art online feature selection methods.
\end{abstract}

\section{Introduction}

High dimensional data present a lot of challenges for data mining and pattern recognition. Fortunately, feature selection is an effective approach to reduce dimensionality by eliminating the irrelevant and redundant features \cite{Guyon2003}. Feature selection efforts can be categorized into two branches: standard feature selection and online feature selection. The former is only performed after all of the features are calculated and obtained, while in some real-world applications such as image analysis and email spam filtering \cite{liu2012online}, features arrive dynamically. It is very time-consuming (if not unrealistic) to wait for all of the incoming features. So it is necessary to perform feature selection incrementally in these applications, which is referred to as online feature selection.

In contrast to standard online learning which supposes the feature space remains consist while the samples flow in sequentially, online feature selection assumes that features flow into the model one by one dynamically and the selection is performed at the time they arrive. It is different from the standard online learning problem which assumes instances arrive dynamically \cite{hoi2012online}. Several online feature selection methods have been proposed recently, such as Grafting \cite{graft}, Alpha-investing \cite{alpha05}, and OSFS \cite{fast2012}. The Grafting approach selects features by minimizing the predefined binomial negative log-likelihood loss function. The Alpha-investing approach evaluates the new feature based on a streamwise regression model. The OSFS approach obtains the optimal subset by the relevance and redundancy analysis. These approaches can evaluate features dynamically with the arrival of each new feature, but they present a common limit: existing online feature selection approaches evaluate the features individually, thus, they overlook the relationship between features which is very important in some real-world data sets. %{\color{blue}
  For example, in image analysis, each image could be represented by multiple kinds of descriptors (groups), such as SIFT %\cite{sift99}
  for shape information and Color Moment %\cite{colormoment}
  for color information, each of which has a high dimension of features. To solve this problem, some researchers have studied the group structure information, such as group Lasso \cite{wang2008note}. However, theses methods limit their applications in online features selection since they require a global feature space in advance.

To overcome the weakness of online feature selection approaches and the limitations of group selection approaches mentioned above, we propose a new online approach for group feature selection in a dynamic feature stream, called Online Group Feature Selection (OGFS). As a group of features arrives, we first introduce our criteria based on spectral analysis to select features with discriminative ability, referred to as online intra-group selection. Second, we refine the sparse linear regression model of Lasso to find the global optimal subset after seeing all features in the group, referred to as online inter-group selection.

With the above motivation, we formulate the problem of online group feature selection with a feature stream and propose a solution to this problem in this paper. Our major contributions can be summarized as follows:
\begin{itemize}
\item Different from the traditional online feature selection, we address the problem of online group feature selection instead of online individual feature selection. To the best of our knowledge, this is the first effort that considers the online group feature selection problem.

\item  To evaluate the features dynamically, we introduce spectral analysis into the online feature selection and provide two novel effective criteria.

\item Our proposed Online Group Feature Selection (OGFS) method achieves the best classification accuracy compared with other state-of-the-art methods for online feature selection.
\end{itemize}

The rest of the paper is organized as follows. Section 2 presents a  review of related work. Section 3 introduces our framework for online group feature selection. Section 4 reports some experimental results to demonstrate the effectiveness of the proposed method. We conclude this work in Section 5.

\section{Related Work}

There are two categories of feature selection approaches: standard feature selection and online methods. Standard feature selection is performed after all the features are computed. It consists of three categories: filter \cite{farahat2012efficient}, wrapper \cite{filter2001} and embedded methods \cite{lasso}. Filter methods usually explore the intrinsic statistical properties of features.
 Wrapper methods use forward or backward strategies to search the whole candidate subsets and a classifier is directly applied.
  Embedded methods attempt to find an optimal subset based on a regression model with specific penalties on coefficients of features. In general, standard feature selection methods process the features individually. Considering some specific applications where the feature space comes with prior knowledge of group structures, some standard methods have been developed accordingly, such as group Lasso.

All of the standard methods mentioned above require the global feature space in advance to perform the selection. However, in some real-world applications, it is difficult to get the global feature space. To overcome this problem of standard feature selection, online feature selection has attracted a lot of attention in recent years. It assumes that features flow in one by one and it aims to discard or select the newly generated feature dynamically. Representative online feature selection approaches include Grafting, Alpha-investing and OSFS. Grafting performs the selection based on a gradient descent technique which has been proven to be effective in pixel classification.
 It still requires a global feature space to define the key parameters in the selection of new features. Hence, it cannot handle the case that the feature stream is infinite or with an unknown size. Alpha-investing evaluates the new feature with a $p$-value returned by a regression model. If the $p$-value of the new feature reaches a certain threshold $\alpha$, the feature will be selected. In Alpha-investing, once the features are selected, they will never be discarded. OSFS selects features based on an online relevance and redundancy analysis. According to the relevance to the class label, input features could be characterized as strongly relevant, weakly relevant or irrelevant. Relevant features will be obtained by online relevance analysis, and redundant features will be removed by Markov blankets. In OSFS, each time when a new feature is included, the redundancy of all selected features will be reanalyzed. To speed up the redundancy analysis, a faster version of OSFS, called Fast-OSFS, was proposed \cite{fast2012}. Fast-OSFS first analyzes the redundancy of a new relevant feature, then decides whether the redundancy analysis of the selected feature subset will be performed or not. It is still inefficient with the increase of selected features. In addition, all of these online feature selection methods evaluate features individually, thus, they overlook the prior knowledge of group information of features.

In contrast to the above existing efforts, we address the online group feature selection problem in this paper. To make use of the prior knowledge of group information, we propose an efficient online feature selection framework including the intra-group feature selection and inter-group feature selection, and based on this framework, we develop a novel algorithm called OGFS.

\section{A Framework for Online Group Feature Selection}

We first formalize our problem for online group feature selection.
Assume a data matrix $X=[x_1,\cdots,x_n]\in \mathbb{R}^{d \times n}$, where $d$ is the number of features arrived so far and $n$ is the number of data points, and a class label vector $Y=[y_1,\cdots,y_n]^{T}\in \mathbb{R}^{n}$, $y_i\in\{1,\cdots,c\}$, where $c$ is the number of classes. The feature space is a dynamic stream vector $F$ consisting of groups of features, $F=[G_1,\cdots,G_j,\cdots]^{T}\in \mathbb{R}^{\sum d_{j}}$, where $d_{j}$ is the number of features in group $G_{j}$. $G_j=[f_{j1},f_{j2},\cdots,f_{jm}]^{T}\in \mathbb{R}^{m}$ where $f_{jk}$ is an individual feature.
In terms of feature stream $F$ and class label vector $Y$, we aim to select an optimal feature subset $U=[g_1,\cdots,g_j,\cdots, g_u]^{T}\in \mathbb{R}^{\sum u_{j}}$ when the algorithm terminates, where $u_{j}$ is the number of groups arrived so far, where $g_j\in \mathbb{R}^{m_g}$, $g_j\subseteq G_j$, $m_g<m$, and $g_j$ can be empty.

To solve this problem, we propose a framework for online group feature selection which has two components: intra-group selection and inter-group selection. The intra-group selection is to process features dynamically at their arrival. That is, when a group of features $G_j$ arrives, we get a subset $G_j'$ from $G_j$. %Existing online feature selection methods could be applied here. However, to evaluate the features more effectively with global group information,
In this part, we design two novel criteria based on spectral analysis to obtain the subset. In terms of the features obtained by the intra-group selection, we further consider the global group information by introducing Lasso to get an optimal subset $g_j$ from $G_j'$, namely the inter-group selection. In the following subsections, we will provide details for intra-group selection and inter-group selection.

 \subsection{Online Intra-Group Selection}

Spectral feature selection methods have demonstrated excellent performance \cite{redundancy}. Given a data matrix $X\in \mathbb{R}^{d\times n}$, a weighted graph with edges between data points close to each other is constructed. Let $S_b\in \mathbb{R}^{n\times n}$ evaluate the between-class distance, and $S_w\in \mathbb{R}^{n\times n}$ evaluate the within-class distances.
In this work, we only consider supervised online feature selection. The between-class affinity matrix $S_b$ and the within-class affinity matrix $S_b$ are calculated as follows \cite{trace2008}:
\begin{equation}\label{sb}
(S_{b})_{ij}=
\begin{cases}
\frac{1}{n}-\frac{1}{n_i},\quad& y_i=y_j, \\
\frac{1}{n}, \quad	 \quad & y_i\neq y_j.
\end{cases}
\end{equation}
\begin{equation}\label{sw}
(S_{w})_{ij}
\begin{cases}
\frac{1}{n_i},\quad& y_i=y_j, \\
0, \quad	 \quad & y_i\neq y_j.
\end{cases}
\end{equation}
where $n_i$ denotes the number of data points from class $i$.

Let the feature selector matrix $W=[w_i,\cdots,w_m]^{T}\in \mathbb{R}^{d\times m}$ where $d$ is the number of features arrived and $m$ is the number of features selected so far. $w_i=[w_{i1},\cdots,w_{id}]^{T}\in \mathbb{R}^{d}$ , where $w_{ij}=1$ indicates that the $j$-th feature is selected, while $w_{ij}=0$ indicates that the $j$-th feature is discarded. Spectral feature selection approaches can be categorized into subset-level selection and feature-level selection approaches. The subset-level selection is to find an optimal subset $U$ by maximizing the following criterion:
\begin{equation}\label{max}
      F(U)=\frac{\text{tr}(W_U^T(XL_bX^T)W_U)}{\text{tr}(W_U^T(XL_wX^T)W_U)},
\end{equation}
where $W_U$ corresponds to the features in subset $U$, $L_{b}$ and $L_{w}$ are the Laplacian matrices, %\cite{chung1997spectral}
$L_{b}=D_{b}-S_{b}$, $D_{b}$ is a diagonal matrix, $D_{b}=\text{diag}(S_{b}\bm{1})$; $L_{w}=D_{w}-S_{w}$, $D_{w}$ is a diagonal matrix and $D_{w}=\text{diag}(S_{w}\bm{1})$.

The feature-level spectral feature selection approach evaluates feature $f_i$ by a score defined below:
\begin{equation}\label{fs}
      s(f_i)= \frac{w_i^T(XL_bX^T)w_i}{w_i^T(XL_wX^T)w_i}.
\end{equation}

After calculating scores of all features, the feature-level approach will select the leading features by the rankings of scores. As traditional spectral feature selection approaches rely on the global information, they are not efficient with online feature selection. Hence, we design two novel spectral-based criteria as follows.

\newtheorem{meth}{Criterion }
\begin{meth}
Given $U\in \mathbb{R}^{b}$ as the previously selected subset, $f_i$ denotes the newly arrived feature, we assume that with the inclusion of a "good" feature, the between-class distances will be maximized, while the within-class distance will be minimized. That is, feature $f_i$ will be selected if the following criterion is satisfied:
\end{meth}
\begin{equation}\label{dis}
 %\begin{split}
     F(U\bigcup f_i)-F(U)>\varepsilon\\
   % \text{s.t.} \  &\varepsilon<0.001\\
%\end{split}
\end{equation}
where $\varepsilon$ is a small positive parameter (we use  $\varepsilon=0.001$ in our experiments.

However, with the increase of selected features, the criterion defined in Eq. (\ref{max}) will be more and more difficult to be satisfied. Hence, to avoid leaving out discriminative features, we design a second criterion.

\begin{meth}
Given $U\in \mathbb{R}^{b}$ as the previously selected subset, and the newly arrived feature $f_i$, we calculate the score of feature $f_i$ by Eq. (\ref{fs}) which shows the discriminative power of the feature. If it is a significant feature with discriminative power, it will be selected.
\end{meth}

The significance of a feature can be evaluated by the $t$-test \cite{ttest} defined bellow:
\begin{equation}\label{test}
      t(f_i,U)= \frac{\hat{\mu}-s(f_i)}{\hat{\sigma}/\sqrt{|U|}}
\end{equation}
where $|U|$ stands for the number of features in $U$, $\hat{\mu}$ and $\hat{\sigma}$ are the mean and standard deviation of scores of all the features in $U$. If the $t$-value returned by Eq. (6) reaches 0.05, then the feature is assumed to be significant among the selected subset $U$ and will be selected ($0.05$ is often used to measure the significance level).

After intra-group selection, we will obtain a subset $G_j'\in \mathbb{R}^{m'}$ from the original feature space $G_j$, $G_j'\subset G_j$. As intra-group selection evaluates the features individually and does not consider the group information, we will apply inter-group selection.

 \subsection{Online Inter-Group Selection}

In this subsection, we introduce Lasso to obtain an optimal subset based on global group information. Given the subset selected during the first phase $G_j'=[f_{j1},f_{j2},\cdots,f_{jm'}]^{T}\in \mathbb{R}^{m'}$, the previously selected subset of features $U^{T}\in \mathbb{R}^{b}$, the combined feature space with dimensionality of $m''$ ($m'+d = m''$), a data set matrix $X \in \mathbb{R}^{m''\times n}$, and a class label vector $Y\in \mathbb{R}^{n}$, $\hat{\beta}=[\hat{\beta}_1,\cdots,\hat{\beta}_{m''}]\in \mathbb{R}^{m''}$ is the projection vector which constructs the predictive variable $ \hat{Y}$:
 \begin{equation}\label{linear}
  \hat{Y} = X^{T}\hat{\beta}    \\
\end{equation}
Lasso chooses an optimal $\hat{\beta}$ by minimizing the objective function defined as follows:
\begin{equation}\label{lasso}
    \begin{split}
   & \min \  ||Y-\hat{Y}||_2^2   \\
    \text{s.t.} & \   ||\beta||_1\leq\lambda, \hat{Y} =X^{T}\hat{\beta}.
      \end{split}
\end{equation}
where $||\cdot||_2$ stands for $l_2$ norm, and $||\cdot||_1$ stands for $l_1$ norm of a vector, $\lambda$ is a parameter that controls the amount of regularization applied to estimators, and $\lambda\geq0$ \cite{lu2012robust}. %If $\lambda$ is set to be a large value, the problem in Eq. (\ref{lasso}) can be solved by Least Squares which minimizes the unregularized square loss. On the other hand, if $\lambda$ is set to be a small value, more entries in $\beta$ will shrink to $0$.
In general, a smaller $\lambda$ will lead to a sparser model. By regression, the component in $\beta_i$  will be set to zero corresponding to feature $f_i$ which is irrelevant to the class label. Finally, the features corresponding to non-zero coefficients will be selected.
After inter-group selection, we get the ultimate subset $U_j$.

 With the combination of the online intra-group and the inter-group selection, the algorithm of Online Group Feature Selection (OGFS for short) can be formed.

 \subsection{OGFS: Online Group Feature Selection Algorithm}

Algorithm 1 shows the pseudo-code of our online group feature selection (OGFS) algorithm. OGFS is divided into two parts: intra-group selection (Steps 4-15) and inter-group selection (Step 16). Details are as follows.

In the intra-group selection, for each feature $f_i$ in group $G_j$, we evaluate features by the criteria defined in Section 3.1. Steps (9-11) evaluate the significance of features based on criterion 1. With the inclusion of the new feature $f_i$, if the within-class distance is minimized and the between-class distance is maximized, feature $f_i$ is thought to be a ``good'' feature and will be added to $G_j'$. Steps (12-14) evaluate the features according to criterion 2.
Based on the selected subset $U$, we validate the significance of the feature by $t$-test. If the $t$-value returned by Eq. (6) is larger than 0.05, feature $f_i$ is thought to be significant in discrimination. Then $f_i$ will be added to $G_j'$.
After intra-group selection, we get a subset of features $G_j'$. To implement the global information of groups, we build a sparse representation model based on the selected subset $U$ and the newly selected subset $G_j'$. An optimal subset $g_j$ will be returned by the objective function defined in Eq. (\ref{lasso}).

In our algorithm, the selected features will be revaluated in the intra-group selection in each iteration. The time complexity of intra-group selection is $O(m)$, and the time complexity of inter-group selection is $O(q)$. Thus, our OGFS algorithm, whose time complexity is linear with the number of features and the number of groups, is very fast.

The iterations will continue until the performance of $\psi(U)$ reaches a predefined threshold as follows:
 \begin{itemize}
 \item $|U|\geq k$, $k$ is the number of features we need to select;
 \item $accu(U)\geq max$, the predictive accuracy of the model based on $U$ reaches the predefined accuracy $max$;
 \item There are no more features to come.
\end{itemize}
{
\renewcommand\algorithmicrequire{\textbf{Input:}}
\renewcommand\algorithmicensure {\textbf{Output:}}
\begin{algorithm}[!t]
\caption{OGFS (Online Group Feature Selection)}
\begin{algorithmic}[1]
\REQUIRE  feature stream $F\in \mathbb{R}^{m*q}$, label vector $Y\in \mathbb{R}^{n}$.
\ENSURE selected subset $U$.
\STATE{ $U=$[], $i=1$, $j=1$;}
\STATE {\textbf{while} $\psi(U)$ not satisfied \textbf{do}}
\FOR{$j=1$ to $q$}
\STATE $G_j \leftarrow$ generate a new group of features;
\FOR{$i=1$ to $m$}
\STATE $G_j' =$ [];
\STATE $f_i\leftarrow$ new feature;
\STATE /*evaluate feature $f_i$ by criterion 1, 2*/
\IF  {$ F(f_i\bigcup G_j')- F(G_j')>\varepsilon $}
\STATE  $G_j' = G_j'\bigcup f_i$;
%\STATE  where $F(f_i\bigcup G_j^{'})$, $F(G_j^{'})$ are defined in Eq. (\ref{max});
\ENDIF
\IF  {$ t(f_i, U) > 0.05 $}
\STATE  $G_j' = G_j'\bigcup f_i$;
%\STATE  where $t(f_i,U)$ is defined in Eq. (6);
\ENDIF
\ENDFOR
\STATE {$g_j\leftarrow$ find the global optimal subset $G_j'$ by Eq. (8)};
\STATE {$U = U \bigcup g_j$}
\ENDFOR
\STATE {\textbf{end while}}
\end{algorithmic}
%\textbf{end while}
\end{algorithm}
}

\section{Experiments}

In this section, extensive experiments are performed to validate the efficiency of our proposed method. We use the benchmark data sets with self-defined group feature structure and two image data sets with pre-existing feature structures. Several state-of-the-art online feature selection methods are used for comparison, including Alpha-investing and Fast-OSFS. The classification accuracy and the compactness (the number of selected features) are used to measure the performances of the algorithms in our experiments.

We divide this section into three subsections, including an introduction to our data sets, the experimental setting in our experiments and the experimental comparison conducted on the benchmark and real-world data sets. Details are as follows.

\subsection{Data Sets}

Our experimental data include benchmark data sets (the first 8 data sets) and real-world data sets (Soccer, the Flower-17 and 15 Scenes, PASCAL VOC 2010) described in Table 1. The column "groups" denotes the number of groups. The eight benchmark data sets are from the UCI repository %\footnote{http://archive.ics.uci.edu/ml} %\cite{UCI}
(the first 4 data sets) and microarray domains\footnote{http://www.cs.binghamton.edu/~lyu/KDD08/data/} (colon, prostate, leukemia, and lungcancer) .
The real-world data sets include: PASCAL VOC 2010 \cite{pascal-voc-2010}, 15 Scenes\footnote{http://www-cvr.ai.uiuc.edu/ponce\_grp/data/}, the Soccer data set\footnote{http://lear.inrialpes.fr/people/vandeweijer/soccer/soccerdata.tar} and the Flower-17 data set\footnote{http://www.robots.ox.ac.uk/ ¡«vgg/data/flowers/}.

The PASCAL VOC 2010 data set consists of 10,103 images from 20 classes. These images range between indoor and outdoor scenes, close-ups and landscapes, and strange viewpoints. The data set is an extremely challenging one because all the images are daily photos obtain from Flicker where the size, viewing angle, illumination, etc appearances of objects and their poses vary significantly, with frequent occlusions .
%\begin{figure*}[!t]
%\centering
%\includegraphics[width=1\textwidth]{voc_photo_cut.pdf}
%\caption{Example images from Pascal VOC 2010 dataset. A $\ast$ in the class means that our method outperforms others}
%\label{fig_uci}
%\end{figure*}

  The 15-Scenes data set contains totally 4485 images from 15 categories,  with the number of images ranging from 200 to 400 per class. We take 100 images per class for training and the rest for testing. In our experiment setup, we use the SPM (Spatial Pyramid Matching) to partition each image into 21 segmentations and extract local information for each patch by the SIFT descriptor. Then the sparse coding is used for vector quantification \cite{zhao2012}.
The Soccer data set contains 280 images from 7 football teams. We take 28 images per class for training and use the rest for testing.
The Flower-17 data set contains 17 categories of flowers. We take 680 images for training and 340 images for testing. For both the Soccer and Flower-17 data sets, we use three descriptors, including PHOG, %\cite{phog},
Color Moment %\cite{colormoment}
and texture.% \cite{texture}.

\begin{table*}[!t]
\centering
\caption{Description of the 11 Data Sets}
\label{Tab_resulttab2}
\centering
\begin{tabular}{|c| c| c| c| c|c|}
\hline
Data Set	&$\#$classes	&\multicolumn{2}{c|}{$\#$instances} & $\#$dim.	& $\#$groups \\ \hline
Wdbc	&2	&\multicolumn{2}{c|}{569} & 31	& - \\ \hline
Ionosphere	&2	&\multicolumn{2}{c|}{351}	&34	&-\\ \hline
Spectf	&2	&\multicolumn{2}{c|}{267 }&	44&	-\\ \hline
Spambase	&2&\multicolumn{2}{c|}{ 4,601}		&57&	-\\ \hline
Colon	&2	&\multicolumn{2}{c|}{2,000 }	&62&	-\\ \hline
Prostate	&2	&\multicolumn{2}{c|}{ 102}	&6,033	&-\\ \hline
Leukemia	&2&	\multicolumn{2}{c|}{72}	&7,129	&-\\ \hline
Lungcancer	&2	&\multicolumn{2}{c|}{ 181}	&12,533	&-\\ \hline
\multirow{2}*{Soccer}&	\multirow{2}*{7} &	$\#$train	& $\#$test	&\multirow{2}*{182 }  &\multirow{2}*{3}	\\ \cline{3-4}
& & 196 & 84 & & \\ \hline
Flower-17&	17&	680	&340	&182	&3 \\ \hline
15 Scenes&	15&	1500	&2,985	&21,504	&21 \\ \hline
\end{tabular}
\end{table*}

\begin{table*}[!t] %(table*)
\centering
\caption{Experimental results on benchmark data sets by (a) Alpha-investing, (b) Fast-OSFS, (c) Baseline, and (d) OGFS.}
\label{Tab_resulttab}
\centering
\begin{tabular}{|c| c| c| c| c| c |c| c |c|}
\hline
\multirow{2}*{Data Set} & \multicolumn{2}{c|}{Alpha-investing} & \multicolumn{2}{c|}{Fast-OSFS} & \multicolumn{2}{c|}{Baseline }& \multicolumn{2}{c|}{OGFS}\\
\cline{2-9} &$\#$dim.	&accu.	&$\#$dim.	&accu.	&$\#$dim.	&accu. &$\#$dim.	&accu.\\ \hline
Wdbc	&19&	0.95	&$\bm{11}$	&0.94 &31	&0.95	&19	&$\bm{0.96}$  \\ \hline
Ionosphere	&$\bm{8}$	&0.90&	9&	0.93 &34	&0.92&	13	&$\bm{0.94}$ \\ \hline
Spectf	&5	&0.75&$\bm{4}$	&0.79   &44	&0.81&	23&$\bm{0.82}$\\ \hline
Spambase	&44	&0.93&	84	&$\bm{0.94}$ &57	&$\bm{0.94}$&$\bm{27}$	&0.93 \\ \hline
Colon	&$\bm{4}$	&0.80 &$\bm{4}$	&2,000	&0.84 &0.86&	49	&$\bm{0.91}$    \\ \hline
Prostate	&$\bm{2}$	&0.89&	5	&0.91&6,033	&0.90&	82	&$\bm{0.98}$  \\ \hline
Leukemia	&$\bm{1}$	&0.65	&5	&0.95	&7,129	&0.95 &52	&$\bm{1.0}$  \\ \hline
Lungcancer	&10	&0.95	&$\bm{7}$	&0.98	&12,533	&0.97&93	&$\bm{0.99}$  \\ \hline
\end{tabular}
\end{table*}%(table*)µ¥À¸

\subsection{Experimental Settings}

We describe the experimental setting here. The threshold parameter $\alpha$ is set to be 0.5 and 0.05 in Alpha-investing and Fast-OSFS, respectively. The sparse linear regression model of Lasso used in the inter-group selection is solved by SPAMS \footnote{http://spams-devel.gforge.inria.fr/} with the parameter $\lambda\in [0.01, 0.5]$.% set to $0.3$.

To simulate online group feature selection, we allow the features to flow in by groups. For the eight benchmark data sets, we define the group structures of the feature space by dividing the feature space of each data set as follows. The global feature stream is represented by $F = [G_1, \cdots, G_i, \cdots]$, where $G_i=[f_{(i-1)*m+1}, f_{(i-1)*m+2}, \cdots, f_{i*m}]$ with $m$ features. In our experiments, we can get optimal results if we set $m\in[5,10]$. %The following experimental results are obtained in the case with \emph{m} = 10.

For the three real-world data sets, we use pre-existing feature groups, each of which represents a descriptor.
That is, for the 15 Scenes data set, the global feature stream $F=[G_1, \cdots, G_{21}]^{T}\in \mathbb{R}^{21*1024}$, where $G_i\in \mathbb{R}^{1024}$ denotes the SIFT descriptor for a local region of the image.
  For the Soccer and Flower-17 data sets, the global feature stream $F=[G_1, G_2, G_3]^{T}\in \mathbb{R}^{182}$, where $G_1\in \mathbb{R}^{168}$ denotes the PHOG descriptor, $G_2\in \mathbb{R}^{6}$ denotes the Color Moment descriptor, and $G_3\in \mathbb{R}^{8}$ denotes the texture descriptor.

The classification of the eight benchmark data sets is based on three classifiers, $k$-NN, J48 and Randomforest in Spider Toolbox\footnote{http://www.kyb.mpg.de/bs/people/spider/main.html}. We adopt 10-fold cross-validation on three classifiers and choose the best accuracy as the final result. For the real-world data sets, we use the nearest neighbor classifier.

All experiments are conducted on a PC computer with Windows XP, 2.5GHz CPU and 2GB memory.

\begin{table*}[!t]
\centering
\caption{Experimental results on real-world data sets by (a) Alpha-investing, (b) Fast-OSFS, (c) Baseline, and (d) OGFS.}
\label{Tab_resulttab3}
\centering
\begin{tabular}{|c| c| c| c| c| c| c|}
\hline
\multirow{2}*{} & \multicolumn{2}{c|}{Soccer} & \multicolumn{2}{c|}{Flower-17} &\multicolumn{2}{c|}{15 Scenes}\\
\cline{2-7} &$\#$dim.	&accu.	&$\#$dim.	&accu.	&$\#$dim.	&accu. \\ \hline
Alpha-investing &	8	&0.25	&$\bm{19}$&	0.329 &$\bm{72}$&	0.393\\ \hline
Fast-OSFS &$\bm{7}$	&0.345	&41	&0.344        &-&	-\\ \hline
Baseline	&182	&0.25	&182	&$\bm{0.347}$ &21,504&	$\bm{0.654}$ \\ \hline
OGFS	&13	&$\bm{0.369}$ &29 &0.344    &369&	0.54\\ \hline
\end{tabular}
\end{table*}

\subsection{Experimental Results on Benchmark Data}
Table 2 shows experimental results of classification accuracy versus compactness on the eight benchmark data sets.
\begin{itemize}

\item OGFS vs. the Baseline%Three feature selection methods vs. Baseline

Though the Baseline is based on the global feature space, our algorithm outperforms Baseline on 7 out of the 8 data sets on both accuracy and compactness. On the data set Spambase, our OGFS is only $1\%$ lower than Baseline but is much more compact. The results show that OGFS could efficiently select the features with most discriminative power.
%All of three feature selection methods will select much smaller number of features from the global feature space while baseline algorithm will maintain the global feature space. In this case, Baseline is superior to Alpha-investing on all the data sets in accuracy, it wins 4 times and ties once in accuracy compared to Fast-OSFS, but it loses 7 out of 8 times compared to our OGFS algorithm. The result shows the discriminative ability of the subset selected by our algorithm.

 \item OGFS vs. Alpha-investing

Alpha-investing obtains more compactness than our OGFS algorithm on 6 data sets, but it loses on 7 out of the 8 data sets in the accuracy except on the Spambase data set. On the Spambase data set, our algorithm achieves the same accuracy as Alpha-investing while obtaining more compactness. More specifically, on data sets Colon and Leukemia, the accuracies of Alpha-investing are 0.80 and 0.65 while OGFS reaches up to 0.91 and 1.0. This is because the previously selected subset will never be revaluated in Alpha-investing, which affects the selection of the later arrived features. However, in our algorithm, %we consider the correlations of the features within a group and the relationship between the groups.
 selected features will be revaluated in the inter-group selection in each iteration. Thus, our algorithm tends to select sufficient features with discriminative power.

\item OGFS vs. Fast-OSFS

Fast-OSFS obtains more compactness on most of the data sets, but our algorithm is better than Fast-OSFS in accuracy on 7 out of the 8 data sets with a little compactness loss. More precisely, on the Spambase data set, the accuracy of our algorithm is slightly lower than Fast-OSFS, but it selects many fewer features. The reason is that Fast-OSFS evaluates features individually rather than in groups. Meanwhile, contrary to Fast-OSFS, our algorithm facilitates the relationship of features within groups and the correlation between groups, which will lead to the optimum of the ultimate subset.
\end{itemize}

Experimental results on benchmark data sets show that our algorithm is superior to Alpha-investing and Fast-OSFS in classification accuracy in most cases, while maintaining the compactness.

\subsection{Experimental Results on Real-world Data Sets}
The results obtained on real-world data sets with pre-existing group structures are shown in Table 3. Since the Fast-OSFS can not perform the classification when the dimensionality is higher than 20,000 (the 15 Scenes data set), as suggested by the author \cite{fast2012}, there is no Fast-OSFS (provided by the authors) is out of memory when performing on this data set. We have the following observations:

\begin{itemize}
\item OGFS vs. the Baseline% Three feature selection methods vs. Baseline

OGFS obtains more compactness than Baseline on all the three data sets. On the Soccer data set, our algorithm obtains the best accuracy. The accuracy of OGFS on Flower-17 is only slightly lower than Baseline. Baseline outperforms OGFS on the 15 Scenes data set, but OGFS selects many fewer features while obtains the best accuracy among all the competing feature selection methods.
 %data set  Alpha-investing achieves the same accuracy as Baseline, and Fast-OSFS obtains a higher accuracy than Baseline with the smallest number of features. On the data set Flower-17, OGFS and Fast-OSFS obtain the same accuracy which is slightly lower than Baseline. Alpha-investing selects the fewest number of features, while loses in accuracy. On the data set 15 Scenes, Alpha-investing achieves good compactness with a a much lower accuracy than the Baseline. The accuracy of OGFS is also lower to Baseline.

\item OGFS vs. Alpha-investing

Compared to Alpha-investing, OGFS obtains higher accuracies with a little compactness loss. In particular, on the 15 Scenes data set, the accuracy of Alpha-investing is only 0.393, while our method could reach 0.54. The reason is the same as we mentioned above. %Besides, on real-world data sets with pre-existing features, the correlation among features is more important, our algorithm could use the information to find the optimal subset.

\item OGFS vs. Fast-OSFS

OGFS outperforms Fast-OSFS on both compactness and accuracy on the Soccer data set. On the Flower-17 data set, OGFS and Fast-OSFS obtain the same accuracy while our algorithm achieves more compactness. These results demonstrate that our algorithm is better than Fast-OSFS when applied on real-world data sets with pre-existing group structures. The reason is the same as we analyzed before.

\end{itemize}

In sum, the above experimental results on real-world data sets reveal the effectiveness of our algorithm, and indicate that our algorithm is more suitable for real-world applications than existing state-of-the-art online feature selection methods.

\section{Conclusion}

In this paper, we have formulated the problem online group feature selection with a feature stream and presented an algorithm called OGFS for this problem. In contrast with traditional online feature selection, we have considered the situation that features arrive by groups in real-world applications. We divided online group feature selection into two stages, i.e., online intra-group and inter-group selection. Then we designed two novel criteria based on spectral analysis for intra-group selection, and introduced Lasso to reduce the redundancy in inter-group selection. Extensive experimental results on benchmark and real-world data sets have demonstrated that OGFS is superior to other state-of-the-art online feature selection methods.
%
%\section*{Acknowledgments}
%
%The authors would like to thank Dr. Canyi Lu and Dr. Peipei Li for their helpful and informative discussions. This research was supported by the National Natural Science Foundation of China (Nos. 61005007, 61229301, 61272366, 61272540, and 61273292), the US National Science Foundation (NSF CCF-0905337), the 973 Program of China (No. 2013CB329604), the 863 Program of China (No. 2012AA011005), the Hong Kong Scholars Program (No. XJ2012012) and the Research Grant Council of Hong Kong SAR (No. HKBU 210309).

\bibliographystyle{named}
\bibliography{ijcai13}

\end{document}